\documentclass{article} 
\usepackage{nips13submit_e,times}
\usepackage{hyperref}
\usepackage{url}
\usepackage{amsfonts}
\usepackage{amsmath}
\usepackage{bbm}
\usepackage{algorithm}
\usepackage{algorithmicx}
\usepackage{algpseudocode}
\usepackage{graphicx}
\usepackage{pdflscape}
\usepackage{diagbox}

\usepackage{booktabs}
\usepackage{colortbl}
\usepackage{color}
\def\arraystretch{1.2}

\title{Tomographic Reconstruction and Regularisation \\with Search Space Expansion and Total Variation}

\author{
Mohammad Majid al-Rifaie\thanks{Corresponding author.} \hspace{0.3cm}
Tim Blackwell \vspace{5mm} 
\\ 
School of Computing \& Mathematical Sciences, University of Greenwich, London, UK
\\
\small{\texttt{ m.alrifaie@gre.ac.uk }}\\
Department of Computing, Goldsmiths College, University of London, London UK\\
\small{\texttt{ t.blackwell@gold.ac.uk }}
}

\nipsfinalcopy 

\begin{document}
\maketitle

\begin{abstract}
The use of ray projections to reconstruct images is a common technique in medical imaging. Dealing with incomplete data is particularly important when a patient is vulnerable to potentially damaging radiation or is unable to cope with the long scanning time. This paper utilises the reformulation of the problem into an optimisation tasks, followed by using a swarm-based reconstruction from highly undersampled data where particles move in image space in an attempt to minimise the reconstruction error. The process is prone to noise and, in addition to the recently introduced search space expansion technique, a further smoothing process, total variation regularisation, is adapted and investigated. The proposed method is shown to produce lower reproduction errors compared to standard tomographic reconstruction toolbox algorithms as well as one of the leading high-dimensional optimisers on the clinically important Shepp-Logan phantom.
\end{abstract}

\section{Introduction}
\label{sec:introduction}
Tomographic reconstruction (TR), which is the determination of the internal structure of an opaque object from projected images cast by penetrating radiation, is at the heart of all medical imaging procedures (X-Ray CT, PET, MRI, Nuclear Medicine and ultrasound~\cite{Giussani2013}) and has widespread application in various scientific fields, mathematics and industry \cite{Irving1994,shliferstein1977some,carazo1999discrete,batenburg2004reconstruction,carvalho1999binary,hampel2007high,nolet2008breviary,butala2010dynamic,gardner1995geometric}.

As part of the imaging process, a number of projections is acquired which are often insufficient for a unique reconstruction. Given the under-determined nature of the problem, and the random nature of the radiation, the operating characteristics of the detectors and, in medical applications, patient movement, projection data is incomplete and noisy. 
The use of an accurate and fast image reconstruction algorithms with the ability to identify structures from reduced patient radiation dose (by weakening the incident radiation and/or reducing acquisition time) is a significant challenge. Success in this `few-view` scenario would enable procedures that were once prohibited.

There are several reconstruction methods which address various challenges posed in the imaging procedure. Filtered backprojection (FBP), is a standard reconstruction technique, which operates with only a single iteration but is not suitable for few-view imaging \cite{Giussani2013}. With increased computational power, Algebraic Reconstruction Techniques (ART) have recently been in more common use. This class of methods are iterative algorithms based on Kacmarz's method~\cite{shliferstein1977some} and, while capable of dealing with few-view scenarios, the added artefacts caused by overfitting is a known weakness. Furthermore, evidence of their worth in large patient populations is lacking~\cite{geyer2015state}. 

If the data is transformed to a sparse and therefore compressible representation, exact image reconstruction is feasible. Compressed Sensing~(CS), which exploits this principle, shows promise for few-view reconstruction, but is heavily dependant on knowledge of the sparse representation, and on replacing the non-convex optimisation problem 
typically unsolvable by traditional methods, by a solvable convex minimisation~\cite{candes2006stable}. 

Maximum-likelihood expectation maximisation (MLEM) and the more computationally efficient ordered subset expectation maximisation (OSEM) are iterative statistical methods which have been trialled and found to be superior to FBP in some cases e.g.~\cite{van2008comparison,trevisan2020comparison}.

Reconstruction problems can be cast as optimisation tasks which opens up the possibility of using population based methods and other metaheuristics (e.g.~\cite{ouaddah2014improving,jarray2010simulated,cipolla2014island,batenburg2009solving,miklos2014particle,tronicke2012crosshole,wang2015image,hu2011clustering}).
An ant algorithm  has been developed for binary reconstruction~\cite{alrifaie_2016_part}.
TR problems and particularly, few-view tomography, is underdetermined: there are many solutions with zero reconstruction error but only some of these are medically feasible and representative of the original object. Artefacts and noise typically reduce reconstruction quality. The challenge is therefore to find biologically plausible solutions of low reconstruction error. 

The research in this paper builds on the findings of reference~\cite{alRifaie2023_SSE} which introduced swarm optimisation with search space expansion (SSE). The idea behind SSE is a successive widening of the effective search space; clamping at each search space boundary favours homogeneity and significantly reduces noise. This work introduces a regularisation procedure in order to further smooth swarm reconstructions. A penalty term, proportional to image gradient, is added to the reconstruction error with the effect that the optimiser must minimise both reconstruction quality and pixel heterogeneity. This approach, known as Total Variation regularisation (TV), has previously been shown to aid CT reconstructions with traditional algorithms \cite{tian2011low}.

This paper compares the performance of FBP and ART toolbox reconstruction algorithms, and a gradient descent reconstruction procedure, CGLS, to a number of population based methods (local and global particle swarm optimisation~(PSO), dispersive flies optimisation~(DFO) and differential evolution~(DE)). The best performing method is then subject to TR-optimised manifestations of SSE and TV. The results are further compared to a state-of-the art high-dimensional optimiser. The statistical analysis and the demonstrable visual reconstruction demonstrates the ability of the proposed method to reconstruct medically pertinent images from limited views.

\section{Tomography and reconstruction}

The problem in discrete reconstruction is: 
\begin{eqnarray}
	\text{find } 
	x &\in& \{0, 1, \ldots, k-1\}^n, k > 1 \nonumber \\
	\text{ such that } Ax &=& b \label{eq:Ax=b}
\end{eqnarray}
where $b \in \mathbbm{R}^m$ is the vector of detector values. A projection matrix, $A \in \mathbbm{R}_{\ge 0}^{m \times n}$, where $m$ is the total number of projections, and $n$ is the number of pixels in the reconstructed image. We consider the problem as greyscale, where $x$ is a vector of reconstructed pixel values and $k = 256$ i.e. $x \in \{0, 1\ldots, 255\}^n$. 

The original object or the ground truth, denoted $x^*$, also satisfying $Ax^* = b$, is either a phantom (artificial test case) or an imaged subject. Given the under-deterministic nature of the problem, the equation $Ax = b$ cannot be inverted (few view, $m << n$) and there are multiple solutions. 
Suppose $y$ is a trial solution; $y$ is forward projected, 
and a \textit{reconstruction} error, $e_1$, evaluated, 
\begin{align}\label{eq:e1}
	e_1(y) &= || b - Ay ||_2^2.
\end{align}

Low reconstruction error does not imply faithfulness to the original object $x^*$. The distance between $y$ to $x^*$ can be measured by a \textit{reproduction} error, $e_2$, in instances where $x^*$ is known:
\begin{align}\label{eq:e2}
	e_2 &= || y - x^* ||_1 
\end{align}
The reproduction error provides a test of the algorithmic ability to find a feasible reconstruction. 

Libraries such as the Astra toolbox~\cite{ASTRA_van2016fast} provide the forward projection operator $A$ in addition to the standard reconstruction algorithms such as ART, FPB and SIRT. 
In the absence of real-world data, and for algorithm development and comparison, a virtual phantom $x^*$ is designed and $b = Ax^*$ is computed by forward projection. The fitness function for optimisation is therefore defined as $e_1(y) = ||b - Ay||_2^2$ where $Ay$ is computed, for trials $y$, by the toolbox.

Discrete TR problems are translatable into real valued problems suitable for optimisers such as PSO and DE by enabling $y$ to take continuous rather than discrete values, $y \in [0, 255]^n$, therefore, allowing $e_1$ and $e_2$ to accept real inputs. The original object, $x^*$, remains discrete. Final solutions $y$ can be discretised for visualisation purposes if necessary. 

\section{Regularisation}

One of the issues associated with image reconstruction is the presence of salt and pepper noise which impacts the fidelity of the reconstructed image to the ground truth. This work uses two approaches to reduce the noise during the reconstruction process: search space expansion (SSE) and total variation regularisation~(TV). 

\subsection{Search space expansion} 

In SSE, the search space allocated to the optimiser is initially confined to only a portion of the feasible space \cite{alRifaie2023_SSE}. Assuming the complete search space to be $\Xi = [0,255]^n$, the initial space would be defined as $\Xi_1 = [0,\frac{255}{d}]^n$, where $d>1$. Any particle leaving this initially allocated space is clamped to the edges. The initial search space is then progressively expanded to explore previous unseen territories. Expansions from the initial space to the subsequent expanded ones kick-in at preset intervals during the optimisation. By the end of the optimisation process, the optimiser can access the entire search space. 

For empirical trials a series of boxes $\Xi_p = [0, \frac{p}{P} \times 255]^n$ are defined with expansions at equal divisions of the total budget of function evaluations. For example, with five subspaces and a budget of 100,000 function evaluations (FEs), search is conducted in the following boxes: 
\begin{align*}
	\Xi_1 &= [0, \frac{255}{5}]^n, &0 < FE \le 20000 \\
	\Xi_2 &= [0, \frac{2 \times 255}{5}]^n, &20000 < FE \le 40000 \\
	\Xi_3 &= [0, \frac{3 \times 255}{5}]^n, &40000 < FE \le 60000 \\
	\Xi_4 &= [0, \frac{4 \times 255}{5}]^n, &60000 < FE \le 80000 \\
	\Xi_5 &= [0, 255]^n, &80000 < FE \le 100000.\\
\end{align*}

\subsection{Total Variation}

To perform the reconstruction task, the problem is formulated in order to retrieve the unknown vector $y$ based on the projection matrix $A$ and the observation vector $b$.
In inverse problems such as TR, regularisation attempts to circumvent overfitting~\cite{chambolle2004algorithm}. Total Variation (TV) regularisation is a deterministic technique that penalises discontinuities in image processing tasks. 

Eq.~\ref{eq:e1} is augmented with a total variation regularisation term:
\begin{eqnarray}
	e_1^{TV}(y) = ||b-Ay||^2_2+ \mu \ TV(y) \label{eq:TV}
\end{eqnarray}
\noindent where the first term serves as a data fidelity term, ensuring the consistency between the reconstructed image $y$ and the measurement $b$, and the second term is the TV semi-norm. $\mu >0 $ controls the influence of the TV term.  

The TV norm is
\begin{align}
	TV(y) = {\sum}_{ij} \sqrt{ |y_{i+1, j} - y_{i,j}|^2 + |y_{i, j+1} - y_{i, j}|^2}
\end{align}
where the sum is over 2D pixel locations $(i, j)$. Regions of small pixel value gradient $|y_{i+1, j} - y_{i,j}|$ and $|y_{i, j+1} - y_{i, j}|$ will minimise $TV(y)$ and provide homogeneous regions of biologically feasible structure.
It has been shown that the TV enhancement is robust, and able to remove noise and artefacts in the reconstructed image~\cite{sidky2008image}.

\section{Minimalist swarm optimiser and TR} 

Dispersive flies optimisation (DFO) is a slimmed-down particle swarm optimisation (PSO) variant, which is distinguished by the abolition of particle memory. Updates are computed from current, rather than historical, position~\cite{alRifaie2014_DFO}. The exploration and exploitation behaviour of the algorithm is investigated in~\cite{al2021exploration}. 
DFO also implements component-wise particle jumps which have been shown to be beneficial in bare bones PSO~\cite{blackwell2011study}.

The optimisation starts by determining the best overall position $g^{t + 1}$, if unique, and positions of all best ring neighbours in the neighbourhood, $n_i^{t + 1}$ of each particle (barring the current swarm best particle, which is not updated). Position component $d$ of all particles $i$, (other than the swarm best) updates according to
\begin{align} \label{eq:dfo}
	&\text{if } u \sim U(0, 1) < \Delta \notag \\
	&\quad x_{id}^{t + 1} \sim U(X_d) \notag \\
	&\text{else} \notag \\
	&\quad x_{id}^{t + 1} = n_{id}^{t + 1} + \phi u_1(g_d^{t+1} - x_{id}^t)
\end{align}
where $\Delta$ is a predetermined jump probability and $U(X_d)$ is the uniform distribution along axis $d$ of the search space $X$, $u_1 \sim U(0, 1)$ and $\phi \in [0, \sqrt{3}]$. The constraint on $\phi$ is derived from a convergence analysis for stochastic difference equations~\cite{blackwell2011study}. The algorithm employs global and local strategies and has two arbitrary parameters $N$ and $\Delta$; and $\phi$ is invariably set to 1 in published studies.
The algorithm has been applied to a wide range of problems in computer vision, aesthetics measurement and art, optimising food processes, electronics, data science and neuroevolution~\cite{foods_2022,acharya2021pid,aparajeya_evomusart_2019,hooman_2018_deep_dfo,alrifaie_2017_evomus_dfo_symcomplexity,alhakbani2017optimising,alrifaie_2016,bishop_2016_acm_autopoiesis_creat_art}.

\section{Experiments and results}

\begin{figure}[h]
	\newcommand{\myW}{0.16\linewidth} 
	\centering
	\includegraphics[width=\myW]{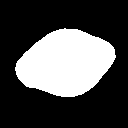}
	\includegraphics[width=\myW]{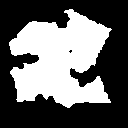}
	\includegraphics[width=\myW]{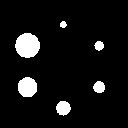}
	\includegraphics[width=\myW]{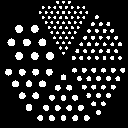}
	\includegraphics[width=\myW]{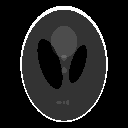}
	\caption{Phantoms}
	\label{fig:phantom}
\end{figure}

The experiments use five phantoms as shown in Fig.~\ref{fig:phantom}. Phantoms 1 - 4 are binary images and phantom 5, the Shepp-Logan phantom, is a discrete problem with six pixel value levels~\cite{shepp1974fourier}. The phantom sizes are $32 \times 32$ and $64 \times 64$. To test few-view, undersampled conditions, the number of projections, $\alpha$, was set to 6, 8, 16 and 32. ASTRA toolbox~\cite{ASTRA_van2016fast} is used to conduct phantom imaging using parallel geometry with the number of rays set to 32 and 64 for the the $32 \times 32$ and $64 \times 64$ phantoms respectively.

From the ASTRA toolbox, five reference algorithms were selected: filtered backprojection (FBP), the algebraic algorithms ART, SIRT and SART, and a gradient descent reconstruction procedure, CGLS. The potential of swarm reconstruction was tested with three swarm algorithms, PSO (in two varieties, global PSO or GPSO, and local PSO, LPSO), DE and DFO.

The swarm algorithms were run for $100,000$ function evaluations. ART, CGLS, FBP, SART and SIRT perform the reconstruction in $\mathbbm{R}^D$, where $D = 32 \times 32 \text{ or } 64 \times 64$. Reconstructions were scaled to $X = [0, 255]^D$ for the purpose of computing the reproduction error, $e_2$. 

A swarm size of $N = 100$ was chosen for G/LPSO, DE and DFO. Particles were initialised in $X$ with the uniform distribution and G/LPSO velocities were set to zero. Particles in all three swarms were clamped to the search box: any particle attempting to leave $X$ was placed on the boundary, and in the case of G/LPSO, velocity is set to zero.

\begin{table} [th]
	
	\caption{Rounded median reconstruction error, $e_1$, for each problem and algorithm.}\label{tab:e1Median}
	\centering \setlength{\tabcolsep}{.5pt} \scriptsize 
	\centering
	\begin{tabular}{|l||c|c|c|c|c||c|c|c|c|}
		\cline{2-10} \cline{3-10} \cline{4-10} \cline{5-10} \cline{6-10} \cline{7-10} \cline{8-10} \cline{9-10} \cline{10-10} 
		\multicolumn{1}{l|}{} & \multicolumn{5}{c||}{\small TR toolbox algorithms} & \multicolumn{4}{c|}{\small Population-based optimisers}\tabularnewline
		\cline{2-10} \cline{3-10} \cline{4-10} \cline{5-10} \cline{6-10} \cline{7-10} \cline{8-10} \cline{9-10} \cline{10-10} 
		\multicolumn{1}{l|}{} & ART  & CGLS  & FBP  & SART  & SIRT  & DE  & DFO  & GPSO  & LPSO \tabularnewline
		\hline 
		\hline
		\textbf{Ph 1}, size = $32^{2}$, $\alpha=6$  & 218  & 16  & 57353136  & 1  & 0  & 6213594  & 6509  & 36756397  & 302287 \tabularnewline
		$\alpha=8$  & 434  & 13  & 55960670  & 9  & 0  & 8972489  & 0  & 51413743  & 33858 \tabularnewline
		$\alpha=16$  & 1192  & 1  & 12080882  & 1421  & 12  & 15729665  & 0  & 104974480  & 4189 \tabularnewline
		$\alpha=32$  & 4767  & 12  & 12993996  & 765224  & 79  & 30193881  & 0  & 202177910  & 2387 \tabularnewline
		\hline 
		Size = $64^{2}$, $\alpha=6$  & 511  & 22  & 881101300  & 2  & 0  & 449457240  & 290703650  & 962231630  & 128747038 \tabularnewline
		$\alpha=8$  & 1993  & 4  & 684892900  & 54  & 1  & 612323100  & 403979650  & 1338255850  & 178845160 \tabularnewline
		$\alpha=16$  & 5347  & 8  & 154832160  & 977  & 12  & 1263333600  & 777572730  & 2558892500  & 343625315 \tabularnewline
		$\alpha=32$  & 32556  & 6  & 95809730  & 52562  & 73  & 2450199350  & 1524608650  & 5098064750  & 666271775 \tabularnewline
		\hline 
		\textbf{Ph 2}, size = $32^{2}$, $\alpha=6$  & 682  & 27  & 219649500  & 0  & 0  & 4881877  & 489974  & 26541697  & 2598464 \tabularnewline
		$\alpha=8$  & 1027  & 0  & 125614740  & 12  & 0  & 7757407  & 290828  & 36228715  & 2010975 \tabularnewline
		$\alpha=16$  & 1014  & 4  & 30864254  & 1144  & 12  & 13824955  & 0  & 75870290  & 163955 \tabularnewline
		$\alpha=32$  & 6805  & 20  & 34511600  & 3798144  & 136  & 27962292  & 0  & 157194035  & 69490 \tabularnewline
		\hline 
		Size = $64^{2}$, $\alpha=6$  & 21532  & 25  & 2552608300  & 7  & 0  & 296865630  & 205105125  & 650832450  & 113333140 \tabularnewline
		$\alpha=8$  & 13219  & 29  & 1476768100  & 121  & 0  & 385430975  & 271130845  & 797472900  & 157673265 \tabularnewline
		$\alpha=16$  & 8634  & 1  & 421734660  & 1859  & 20  & 811162880  & 559093000  & 1715699050  & 324069900 \tabularnewline
		$\alpha=32$  & 50550  & 11  & 278557820  & 42777  & 289  & 1649504250  & 1108368850  & 3431840500  & 634350900 \tabularnewline
		\hline 
		\textbf{Ph 3}, size = $32^{2}$, $\alpha=6$  & 88  & 0  & 118908584  & 0  & 0  & 7825271  & 134  & 67812627  & 736763 \tabularnewline
		$\alpha=8$  & 124  & 0  & 74234100  & 1  & 0  & 10517076  & 56  & 80654290  & 1326800 \tabularnewline
		$\alpha=16$  & 1543  & 0  & 9678624  & 3988  & 13  & 24247643  & 0  & 168775595  & 2079945 \tabularnewline
		$\alpha=32$  & 5894  & 2  & 4692086  & 4138801  & 97  & 42903300  & 0  & 330850820  & 2896024 \tabularnewline
		\hline 
		Size = $64^{2}$, $\alpha=6$  & 3449  & 0  & 1572601300  & 0  & 0  & 835896350  & 503486890  & 2165751550  & 195114935 \tabularnewline
		$\alpha=8$  & 491  & 1  & 1007831400  & 13  & 0  & 1019598350  & 581618945  & 2758263850  & 241910610 \tabularnewline
		$\alpha=16$  & 6827  & 4  & 188459900  & 1472  & 28  & 2082434150  & 1198967450  & 5413350650  & 515041345 \tabularnewline
		$\alpha=32$  & 17844  & 2  & 36140108  & 15064  & 149  & 3993287550  & 2389057650  & 10727333500  & 990994525 \tabularnewline
		\hline 
		\textbf{Ph 4}, size = $32^{2}$, $\alpha=6$  & 217  & 2  & 808024770  & 2  & 0  & 4994230  & 609809  & 25554527  & 3385390 \tabularnewline
		$\alpha=8$  & 555  & 1  & 370298020  & 30  & 1  & 5256030  & 721491  & 24894351  & 4710365 \tabularnewline
		$\alpha=16$  & 8929  & 0  & 69808620  & 10431  & 57  & 23979954  & 2613173  & 72151635  & 14181106 \tabularnewline
		$\alpha=32$  & 47093  & 2  & 26453604  & 4325603  & 590  & 60286648  & 826782  & 157362800  & 27365272 \tabularnewline
		\hline 
		Size = $64^{2}$, $\alpha=6$  & 8068  & 7  & 13681514000  & 10  & 0  & 261069665  & 111867240  & 769442225  & 85870980 \tabularnewline
		$\alpha=8$  & 1008  & 3  & 4972211700  & 1  & 0  & 243611335  & 95635225  & 714078650  & 91989665 \tabularnewline
		$\alpha=16$  & 12288  & 0  & 1202136800  & 1727  & 62  & 582503870  & 259003385  & 1622731400  & 279910750 \tabularnewline
		$\alpha=32$  & 78954  & 1  & 269898500  & 240709  & 702  & 1278379300  & 602676550  & 2860742250  & 706395100 \tabularnewline
		\hline 
		\textbf{Ph 5}, size = $32^{2}$, $\alpha=6$  & 322  & 2  & 64488092  & 5  & 0  & 3358653  & 356519  & 19843914  & 2682284 \tabularnewline
		$\alpha=8$  & 429  & 0  & 37028332  & 102  & 6  & 5220447  & 508443  & 28824148  & 4289071 \tabularnewline
		$\alpha=16$  & 6785  & 2  & 7614852  & 2373  & 29  & 16464641  & 1470355  & 66384160  & 10751817 \tabularnewline
		$\alpha=32$  & 5198  & 2  & 9878102  & 90968  & 74  & 44725869  & 4028948  & 153507290  & 24552911 \tabularnewline
		\hline 
		Size = $64^{2}$, $\alpha=6$  & 2881  & 4  & 1179886800  & 0  & 0  & 240579325  & 98721218  & 821075400  & 69743575 \tabularnewline
		$\alpha=8$  & 1581  & 0  & 663436700  & 218  & 1  & 328983280  & 136800775  & 1059109055  & 107567910 \tabularnewline
		$\alpha=16$  & 5234  & 0  & 159528460  & 890  & 19  & 721549070  & 291544490  & 2089936600  & 248175780 \tabularnewline
		$\alpha=32$  & 37583  & 0  & 62805610  & 58163  & 145  & 1414172750  & 600360370  & 4179300750  & 532845370 \tabularnewline
		\hline 
	\end{tabular}
\end{table}

\begin{table} [h]
	\setlength{\tabcolsep}{2pt}   \small \tiny  \scriptsize
	\renewcommand\arraystretch{1.12}
	\caption{Algorithms comparison based on $e_1$. Numbers indicate statistically significant wins for the algorithm in the left hand column versus the algorithm in the top row.} \label{tab:e1}
	\centering
	\begin{tabular}{|l||c|c|c|c|c||c|c|c|c||c|c|}
		\hline 
		\textbf{ } & ART & CGLS & FBP & SART & SIRT & DE  & DFO  & GPSO  & LPSO  & $\Sigma$ \tabularnewline
		\hline 
		\hline 
		ART & \cellcolor{black!25} NA & 0  & 40  & 12  & 0  & 40  & 31  & 40  & 39  & 202  \tabularnewline 
		\hline  
		CGLS & 40  & \cellcolor{black!25} NA & 40  & 32  & 22  & 40  & 33  & 40  & 40  & 287 \tabularnewline 
		\hline  
		FBP & 0  & 0  & \cellcolor{black!25} NA & 0  & 0  & 16  & 9  & 25  & 10  & 60 \tabularnewline 
		\hline  
		SART & 28  & 8  & 40  & \cellcolor{black!25} NA & 0  & 40  & 32  & 40  & 37 & 225  \tabularnewline 
		\hline  
		SIRT & 40  & 18  & 40  & 40  & \cellcolor{black!25} NA & 40  & 33  & 40  & 40  & \cellcolor{black!20} 291 \tabularnewline 
		\hline  
		DE & 0  & 0  & 23  & 0  & 0  & \cellcolor{black!25} NA & 0  & 40  & 0  & 63 \tabularnewline 
		\hline  
		DFO & 7  & 7  & 31  & 8  & 6  & 40  & \cellcolor{black!25} NA & 40  & 22  & 161 \tabularnewline 
		\hline  
		GPSO & 0  & 0  & 13  & 0  & 0  & 0  & 0  & \cellcolor{black!25} NA & 0  & 13 \tabularnewline 
		\hline  
		LPSO & 1  & 0  & 29  & 3  & 0  & 40  & 18  & 40  & \cellcolor{black!25} NA  & 131 \tabularnewline 
		\hline  
	\end{tabular}
\end{table}

The DFO jump probability was set to $0.001$; G/LPSO was run with $w = 0.729844$ and $c = 1.49618$ and the DE/best/1 parameters $F$ and $C_R$ were both set to $0.5$.
All algorithms with randomisation were run 30 times on each of the 40 problems: 5 phantoms $\times$ 4 projection types (6, 8, 16, 32) $\times$ 2 sizes ($32\times32$ and $64\times64$).

\begin{table} [t]
	\setlength{\tabcolsep}{3pt} \scriptsize 
	\caption{Rounded median reproduction error, $e_2$, for each problem and each algorithm. Lighter shading indicates the proximity of the reconstructions to the phantoms. The largest error in phantoms of sizes $32^2$ and $64^2$ are $255\times 32^2$ and $255\times 64^2$ respectively.}\label{tbl:e2Median}
	
	\centering
	\begin{tabular}{|l||c|c|c|c|c||c|c|c|c|}
		\cline{2-10} \cline{3-10} \cline{4-10} \cline{5-10} \cline{6-10} \cline{7-10} \cline{8-10} \cline{9-10} \cline{10-10} 
		\multicolumn{1}{l|}{} & \multicolumn{5}{c||}{\small TR toolbox algorithms} & \multicolumn{4}{c|}{\small Population-based optimisers}\tabularnewline
		\cline{2-10} \cline{3-10} \cline{4-10} \cline{5-10} \cline{6-10} \cline{7-10} \cline{8-10} \cline{9-10} \cline{10-10} 
		\multicolumn{1}{l|}{} & ART  & CGLS  & FBP  & SART  & SIRT  & DE  & DFO  & GPSO  & LPSO \tabularnewline
		\hline 
		\hline
		\textbf{Ph 1}, size = $32^{2}$, $\alpha=6$  & \cellcolor{black!20}51826  & \cellcolor{black!38}99356  & \cellcolor{black!31}82043  & \cellcolor{black!20}52562  & \cellcolor{black!20}52254  & \cellcolor{black!5}13242  & \cellcolor{black!0}420  & \cellcolor{black!11}29844  & \cellcolor{black!1}2773 \tabularnewline
		$\alpha=8$  & \cellcolor{black!20}51750  & \cellcolor{black!25}66418  & \cellcolor{black!33}86995  & \cellcolor{black!20}52311  & \cellcolor{black!20}53461  & \cellcolor{black!5}11821  & \cellcolor{black!0}1  & \cellcolor{black!11}27672  & \cellcolor{black!0}680 \tabularnewline
		$\alpha=16$  & \cellcolor{black!15}38745  & \cellcolor{black!44}115808  & \cellcolor{black!21}55146  & \cellcolor{black!23}58775  & \cellcolor{black!16}42220  & \cellcolor{black!3}8124  & \cellcolor{black!0}0  & \cellcolor{black!10}25746  & \cellcolor{black!0}150 \tabularnewline
		$\alpha=32$  & \cellcolor{black!14}37587  & \cellcolor{black!3}6821  & \cellcolor{black!13}35016  & \cellcolor{black!43}112567  & \cellcolor{black!11}27634  & \cellcolor{black!3}6962  & \cellcolor{black!0}0  & \cellcolor{black!9}24352  & \cellcolor{black!0}80 \tabularnewline
		\hline 
		Size = $64^{2}$, $\alpha=6$  & \cellcolor{black!21}216081  & \cellcolor{black!21}217040  & \cellcolor{black!41}425778  & \cellcolor{black!21}214884  & \cellcolor{black!24}246053  & \cellcolor{black!15}159530  & \cellcolor{black!14}142480  & \cellcolor{black!21}220776  & \cellcolor{black!9}97210 \tabularnewline
		$\alpha=8$  & \cellcolor{black!22}227658  & \cellcolor{black!27}286941  & \cellcolor{black!37}385363  & \cellcolor{black!22}225173  & \cellcolor{black!23}241075  & \cellcolor{black!15}152623  & \cellcolor{black!13}135958  & \cellcolor{black!21}217264  & \cellcolor{black!9}91322 \tabularnewline
		$\alpha=16$  & \cellcolor{black!16}162459  & \cellcolor{black!20}210787  & \cellcolor{black!25}262494  & \cellcolor{black!19}195464  & \cellcolor{black!17}179756  & \cellcolor{black!15}152407  & \cellcolor{black!12}129482  & \cellcolor{black!20}211900  & \cellcolor{black!8}84812 \tabularnewline
		$\alpha=32$  & \cellcolor{black!13}132566  & \cellcolor{black!12}130101  & \cellcolor{black!17}181438  & \cellcolor{black!44}464631  & \cellcolor{black!14}150986  & \cellcolor{black!14}149321  & \cellcolor{black!12}128086  & \cellcolor{black!20}210921  & \cellcolor{black!8}80886 \tabularnewline
		\hline 
		\textbf{Ph 2}, size = $32^{2}$, $\alpha=6$  & \cellcolor{black!29}75478  & \cellcolor{black!45}117958  & \cellcolor{black!37}96146  & \cellcolor{black!29}74991  & \cellcolor{black!29}75948  & \cellcolor{black!8}20278  & \cellcolor{black!4}9493  & \cellcolor{black!14}35795  & \cellcolor{black!6}16268 \tabularnewline
		$\alpha=8$  & \cellcolor{black!25}66315  & \cellcolor{black!42}110549  & \cellcolor{black!35}91503  & \cellcolor{black!27}70612  & \cellcolor{black!27}71292  & \cellcolor{black!7}18407  & \cellcolor{black!2}5107  & \cellcolor{black!13}33860  & \cellcolor{black!4}9631 \tabularnewline
		$\alpha=16$  & \cellcolor{black!21}53931  & \cellcolor{black!40}104435  & \cellcolor{black!26}67266  & \cellcolor{black!24}62931  & \cellcolor{black!20}52404  & \cellcolor{black!4}11191  & \cellcolor{black!0}0  & \cellcolor{black!10}27210  & \cellcolor{black!1}1327 \tabularnewline
		$\alpha=32$  & \cellcolor{black!14}35972  & \cellcolor{black!6}16688  & \cellcolor{black!17}45266  & \cellcolor{black!46}119778  & \cellcolor{black!11}29999  & \cellcolor{black!3}8957  & \cellcolor{black!0}0  & \cellcolor{black!10}26880  & \cellcolor{black!0}536 \tabularnewline
		\hline 
		Size = $64^{2}$, $\alpha=6$  & \cellcolor{black!29}306143  & \cellcolor{black!47}489520  & \cellcolor{black!38}396555  & \cellcolor{black!29}300664  & \cellcolor{black!30}308809  & \cellcolor{black!17}178848  & \cellcolor{black!16}170864  & \cellcolor{black!22}230096  & \cellcolor{black!13}135870 \tabularnewline
		$\alpha=8$  & \cellcolor{black!26}271574  & \cellcolor{black!44}457891  & \cellcolor{black!39}408102  & \cellcolor{black!26}275737  & \cellcolor{black!27}280536  & \cellcolor{black!17}173193  & \cellcolor{black!16}164081  & \cellcolor{black!22}227429  & \cellcolor{black!12}129423 \tabularnewline
		$\alpha=16$  & \cellcolor{black!22}230965  & \cellcolor{black!22}231018  & \cellcolor{black!31}319252  & \cellcolor{black!24}248662  & \cellcolor{black!23}241230  & \cellcolor{black!15}158693  & \cellcolor{black!14}146497  & \cellcolor{black!21}217395  & \cellcolor{black!10}109135 \tabularnewline
		$\alpha=32$  & \cellcolor{black!17}178735  & \cellcolor{black!17}179516  & \cellcolor{black!21}217124  & \cellcolor{black!37}389843  & \cellcolor{black!17}181413  & \cellcolor{black!15}154848  & \cellcolor{black!13}139559  & \cellcolor{black!21}216685  & \cellcolor{black!10}101305 \tabularnewline
		\hline 
		\textbf{Ph 3}, size = $32^{2}$, $\alpha=6$  & \cellcolor{black!21}54111  & \cellcolor{black!42}110037  & \cellcolor{black!29}75229  & \cellcolor{black!20}53060  & \cellcolor{black!23}61049  & \cellcolor{black!4}11359  & \cellcolor{black!0}34  & \cellcolor{black!12}30476  & \cellcolor{black!1}3542 \tabularnewline
		$\alpha=8$  & \cellcolor{black!26}69160  & \cellcolor{black!30}77480  & \cellcolor{black!33}87131  & \cellcolor{black!27}70097  & \cellcolor{black!25}65759  & \cellcolor{black!5}12164  & \cellcolor{black!0}19  & \cellcolor{black!12}31467  & \cellcolor{black!2}4563 \tabularnewline
		$\alpha=16$  & \cellcolor{black!17}43337  & \cellcolor{black!30}77609  & \cellcolor{black!22}57842  & \cellcolor{black!33}87346  & \cellcolor{black!15}40388  & \cellcolor{black!4}10360  & \cellcolor{black!0}0  & \cellcolor{black!12}32132  & \cellcolor{black!1}2537 \tabularnewline
		$\alpha=32$  & \cellcolor{black!13}33606  & \cellcolor{black!4}9854  & \cellcolor{black!14}36061  & \cellcolor{black!42}110320  & \cellcolor{black!12}31623  & \cellcolor{black!3}8781  & \cellcolor{black!0}0  & \cellcolor{black!12}31635  & \cellcolor{black!1}1756 \tabularnewline
		\hline 
		Size = $64^{2}$, $\alpha=6$  & \cellcolor{black!23}239804  & \cellcolor{black!36}376982  & \cellcolor{black!38}396969  & \cellcolor{black!23}236242  & \cellcolor{black!23}243335  & \cellcolor{black!16}164059  & \cellcolor{black!13}138704  & \cellcolor{black!23}238811  & \cellcolor{black!9}90967 \tabularnewline
		$\alpha=8$  & \cellcolor{black!26}275303  & \cellcolor{black!45}470541  & \cellcolor{black!39}408217  & \cellcolor{black!26}271385  & \cellcolor{black!28}297473  & \cellcolor{black!16}171601  & \cellcolor{black!14}144178  & \cellcolor{black!24}245820  & \cellcolor{black!9}98706 \tabularnewline
		$\alpha=16$  & \cellcolor{black!20}207820  & \cellcolor{black!45}475237  & \cellcolor{black!30}315308  & \cellcolor{black!19}200323  & \cellcolor{black!21}219470  & \cellcolor{black!16}170917  & \cellcolor{black!13}138741  & \cellcolor{black!23}242887  & \cellcolor{black!9}92402 \tabularnewline
		$\alpha=32$  & \cellcolor{black!14}141076  & \cellcolor{black!14}142095  & \cellcolor{black!18}192508  & \cellcolor{black!46}482340  & \cellcolor{black!14}141896  & \cellcolor{black!16}169328  & \cellcolor{black!13}138694  & \cellcolor{black!24}246142  & \cellcolor{black!9}89084 \tabularnewline
		\hline 
		\textbf{Ph 4}, size = $32^{2}$, $\alpha=6$  & \cellcolor{black!41}106233  & \cellcolor{black!41}106734  & \cellcolor{black!41}106053  & \cellcolor{black!41}107896  & \cellcolor{black!41}106042  & \cellcolor{black!16}42639  & \cellcolor{black!14}36845  & \cellcolor{black!20}52692  & \cellcolor{black!16}41449 \tabularnewline
		$\alpha=8$  & \cellcolor{black!38}99030  & \cellcolor{black!48}126284  & \cellcolor{black!43}112713  & \cellcolor{black!38}98011  & \cellcolor{black!39}101802  & \cellcolor{black!19}50782  & \cellcolor{black!17}45347  & \cellcolor{black!22}57539  & \cellcolor{black!19}48984 \tabularnewline
		$\alpha=16$  & \cellcolor{black!30}78535  & \cellcolor{black!42}109904  & \cellcolor{black!36}93446  & \cellcolor{black!45}118392  & \cellcolor{black!35}90560  & \cellcolor{black!15}38058  & \cellcolor{black!9}22937  & \cellcolor{black!19}50469  & \cellcolor{black!13}32753 \tabularnewline
		$\alpha=32$  & \cellcolor{black!24}61504  & \cellcolor{black!7}18723  & \cellcolor{black!29}76074  & \cellcolor{black!57}148488  & \cellcolor{black!23}59498  & \cellcolor{black!12}30381  & \cellcolor{black!1}3817  & \cellcolor{black!18}45841  & \cellcolor{black!9}22629 \tabularnewline
		\hline 
		Size = $64^{2}$, $\alpha=6$  & \cellcolor{black!34}359476  & \cellcolor{black!44}457919  & \cellcolor{black!33}341521  & \cellcolor{black!34}351830  & \cellcolor{black!35}369252  & \cellcolor{black!22}227959  & \cellcolor{black!19}198832  & \cellcolor{black!27}286299  & \cellcolor{black!19}194556 \tabularnewline
		$\alpha=8$  & \cellcolor{black!35}370377  & \cellcolor{black!49}508417  & \cellcolor{black!38}397320  & \cellcolor{black!36}379010  & \cellcolor{black!39}402508  & \cellcolor{black!23}242933  & \cellcolor{black!21}216563  & \cellcolor{black!28}288858  & \cellcolor{black!21}216007 \tabularnewline
		$\alpha=16$  & \cellcolor{black!31}327332  & \cellcolor{black!30}311610  & \cellcolor{black!34}353516  & \cellcolor{black!32}334652  & \cellcolor{black!36}378013  & \cellcolor{black!22}232429  & \cellcolor{black!19}199097  & \cellcolor{black!28}293588  & \cellcolor{black!19}200125 \tabularnewline
		$\alpha=32$  & \cellcolor{black!25}260130  & \cellcolor{black!23}241566  & \cellcolor{black!30}315026  & \cellcolor{black!45}470472  & \cellcolor{black!24}250627  & \cellcolor{black!22}228127  & \cellcolor{black!17}182692  & \cellcolor{black!28}291675  & \cellcolor{black!18}191472 \tabularnewline
		\hline 
		\textbf{Ph 5}, size = $32^{2}$, $\alpha=6$  & \cellcolor{black!31}81618  & \cellcolor{black!48}125584  & \cellcolor{black!33}85029  & \cellcolor{black!32}82844  & \cellcolor{black!31}80749  & \cellcolor{black!16}42046  & \cellcolor{black!14}37052  & \cellcolor{black!20}53284  & \cellcolor{black!16}41565 \tabularnewline
		$\alpha=8$  & \cellcolor{black!26}69074  & \cellcolor{black!27}69497  & \cellcolor{black!36}93444  & \cellcolor{black!27}69635  & \cellcolor{black!27}71007  & \cellcolor{black!16}40768  & \cellcolor{black!13}34377  & \cellcolor{black!20}52579  & \cellcolor{black!15}40438 \tabularnewline
		$\alpha=16$  & \cellcolor{black!17}45410  & \cellcolor{black!37}97915  & \cellcolor{black!25}65674  & \cellcolor{black!31}82172  & \cellcolor{black!18}46194  & \cellcolor{black!14}37515  & \cellcolor{black!11}27831  & \cellcolor{black!19}50481  & \cellcolor{black!13}33405 \tabularnewline
		$\alpha=32$  & \cellcolor{black!10}25400  & \cellcolor{black!1}1332  & \cellcolor{black!17}45441  & \cellcolor{black!35}91940  & \cellcolor{black!6}15897  & \cellcolor{black!13}34474  & \cellcolor{black!8}21385  & \cellcolor{black!18}47786  & \cellcolor{black!11}29791 \tabularnewline
		\hline 
		Size = $64^{2}$, $\alpha=6$  & \cellcolor{black!28}289093  & \cellcolor{black!40}419775  & \cellcolor{black!31}326454  & \cellcolor{black!28}287494  & \cellcolor{black!28}292781  & \cellcolor{black!21}216387  & \cellcolor{black!17}180379  & \cellcolor{black!27}278224  & \cellcolor{black!18}184707 \tabularnewline
		$\alpha=8$  & \cellcolor{black!27}283383  & \cellcolor{black!41}433439  & \cellcolor{black!36}379237  & \cellcolor{black!27}286974  & \cellcolor{black!28}288103  & \cellcolor{black!21}214797  & \cellcolor{black!17}178452  & \cellcolor{black!27}276883  & \cellcolor{black!17}178909 \tabularnewline
		$\alpha=16$  & \cellcolor{black!22}225719  & \cellcolor{black!22}230242  & \cellcolor{black!28}291806  & \cellcolor{black!24}253042  & \cellcolor{black!22}224918  & \cellcolor{black!21}214866  & \cellcolor{black!16}171819  & \cellcolor{black!27}279062  & \cellcolor{black!16}170918 \tabularnewline
		$\alpha=32$  & \cellcolor{black!14}145426  & \cellcolor{black!15}159103  & \cellcolor{black!17}180867  & \cellcolor{black!44}461802  & \cellcolor{black!18}184953  & \cellcolor{black!20}211410  & \cellcolor{black!16}167643  & \cellcolor{black!27}278226  & \cellcolor{black!16}168244 \tabularnewline
		\hline 
	\end{tabular}
\end{table}

\begin{table} 
	\setlength{\tabcolsep}{2pt}  \scriptsize  \small \tiny  \scriptsize
	\caption{Algorithms comparison based on $e_2$. Numbers indicate statistically significant wins for the algorithm in the left hand column versus the algorithm in the top row.}
\label{tab:e2}
\centering
\begin{tabular}{|l||c|c|c|c|c||c|c|c|c||c|c|}
	\hline 
	\textbf{ } & ART & CGLS & FBP & SART & SIRT & DE  & DFO  & GPSO  & LPSO & $\Sigma$ \tabularnewline
	\hline 
	\hline 
	ART & \cellcolor{black!25} NA & 32  & 37  & 29  & 28  & 4  & 1  & 11  & 2  & 144 \tabularnewline 
	\hline  
	CGLS & 8  & \cellcolor{black!25} NA & 20  & 17  & 13  & 5  & 2  & 12  & 3  & 80 \tabularnewline 
	\hline  
	FBP & 3  & 20  & \cellcolor{black!25} NA & 16  & 4  & 1  & 0  & 4  & 0  & 48 \tabularnewline 
	\hline  
	SART & 11  & 23  & 24  & \cellcolor{black!25} NA & 18  & 0  & 0  & 4  & 0  & 80 \tabularnewline 
	\hline  
	SIRT & 12  & 27  & 36  & 22  & \cellcolor{black!25} NA & 3  & 1  & 10  & 1  & 112 \tabularnewline 
	\hline  
	DE & 36  & 34  & 39  & 40  & 37  & \cellcolor{black!25} NA & 0  & 40  & 0  & 226 \tabularnewline 
	\hline  
	DFO & 39  & 38  & 40  & 40  & 39  & 40  & \cellcolor{black!25} NA & 40  & 22  & \cellcolor{black!20} 298 \tabularnewline 
	\hline  
	GPSO & 28  & 27  & 35  & 35  & 28  & 0  & 0  & \cellcolor{black!25} NA & 0  & 153 \tabularnewline 
	\hline  
	LPSO & 38  & 37  & 40  & 40  & 39  & 39  & 13  & 40  & \cellcolor{black!25} NA  & 286 \tabularnewline 
	\hline  
\end{tabular}
\end{table}

\subsection{Vanilla swarm and toolbox algorithms results}

Aggregating the results of Table~\ref{tab:e1Median} which summarises the performance of the swarm and toolbox algorithms, Table~\ref{tab:e1} reports on Wilcoxon statistical significance tests on the reconstruction error for algorithm pairs for the 40 problem instances at a significance level of $0.05$. The rows show the number of instances in which the row algorithm performed better than the column algorithm. For example, reading along the first row, ART gave a significantly smaller reconstruction error, $e_1$, than SART on 12 of the 40 trials.

Algebraic reconstruction algorithms ART, SART and SIRT and gradient descent, CGLS, uniformly outperform DE, GPSO and LPSO and are better than DFO in at least 31 trials. SIRT presents the best performance as shown in shaded colour in the last column which sums the total number of significant wins. These results are expected given the toolbox algorithms have been specifically designed for reconstruction while the swarm algorithms are off-the-shelf multi-purpose low-dimensional optimisers that have not been tuned to the reconstruction task. 

In Table~\ref{tab:e2}, comparing algorithm pairs based on the reproduction error, $e_2$, illustrates that DE, LPSO (and to a lesser extent GPSO) consistently find better reproductions than any of the toolbox algorithms; and DFO produces better results than the rest of the swarm and toolbox algorithms. 

Table~\ref{tbl:e2Median} presents the median reproduction error of each algorithm over each set of problem. Lighter shading indicates lower $e_2$ error. DFO generates good reproduction, followed by LPSO. DFO produces exact reconstruction of the original phantoms ($e_2=0$) in 6 out of 40 experiments, where no other swarm or toolbox algorithm is capable of producing an exact reconstruction.

\subsection{Parameter tuning}

To better optimise the leading swarm method in the context of the problem, DFO parameters were fine-tuned. Phantom $1$, with size $32\times32$ and $\alpha=6$, was used as a benchmark to sweep through the parameter space. The parameters tuned are $N,\phi$ and $\Delta$, and DFO was used as a hyper-parameter optimiser. The optimiser was run 30 times with the population size of $10$ and the termination criterion set to $100$ iterations ($1000$ function evaluations). During the fine-tuning exercise, the elitism mechanism was revised, catering for the presence of stochasticity in the optimisation process, and to enable re-evaluation of the current best particle parameters found in the trials. The optimum and most frequent successful parameter values found were $\Delta=0.001$, $N=2$ and $\phi=1.7320508\approx\sqrt{3}$. The small value of optimal $N$ confirms the finding that the function profile has a single broad funnel leading the global optima, in effect rendering the task into a largely unimodal problem~\cite{alRifaie2022swarmTR}. It also indicates that DFO is acting as a `swarm-inspired local search' and the collective presence of a large communication network is unnecessary in this context. Three local search algorithms (Nelder-Mead~\cite{nelder-mead1965simplex}, L-BFGS-B~\cite{L-BFGS-B_zhu1997algorithm}, MTS-LS1~\cite{MTS_tseng2008multiple}), had been investigated, demonstrating that a `simple' local search may fall short of competing with `a swarm-based local search'.

\subsection{Search space expansion}

The positive impact of SSE has been studied by researchers. In this work, first SSE is fine-tuned on Shepp-Logan phantom, which was introduced in 1974, and is a schematic representation of a cranial slice~\cite{shepp1974fourier} (a standard phantom for TR algorithm testing); then, the fitness function is adapted to integrate the TV regularisation term~(Eq.~\ref{eq:TV}). Fine-tuning SSE identifies the optimal number of expanding boxes required to reduce the noise in the reconstructed images.

As a result, Table~\ref{tab:e1e2SheppLogan} presents the median errors for the reconstruction and reproduction errors, $e_1$ and $e_2$ respectively, for 30 runs. 

\begin{figure}
	\centering
	\newcommand{\myW}{0.072\linewidth} 
	\setlength{\tabcolsep}{1pt}
	\begin{tabular}{cccccccccccc} 
		\fbox{\diagbox[width=4.5em]{No.\\of boxes}{FEs}} & 10k & 20k & 30k & 40k & 50k & 60k & 70k & 80k & 90k & 100k \tabularnewline
		1 &
		\includegraphics[width=\myW]{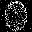} & 
		\includegraphics[width=\myW]{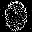} & 
		\includegraphics[width=\myW]{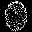} & 
		\includegraphics[width=\myW]{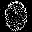} & 
		\includegraphics[width=\myW]{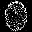} & 
		\includegraphics[width=\myW]{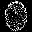} & 
		\includegraphics[width=\myW]{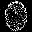} & 
		\includegraphics[width=\myW]{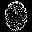} & 
		\includegraphics[width=\myW]{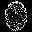} & 
		\includegraphics[width=\myW]{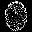}\tabularnewline
		
		2 &
		\includegraphics[width=\myW]{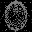} & 
		\includegraphics[width=\myW]{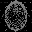} & 
		\includegraphics[width=\myW]{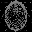} & 
		\includegraphics[width=\myW]{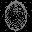} & 
		\includegraphics[width=\myW]{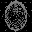} & 
		\includegraphics[width=\myW]{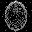} & 
		\includegraphics[width=\myW]{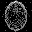} & 
		\includegraphics[width=\myW]{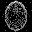} & 
		\includegraphics[width=\myW]{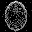} & 
		\includegraphics[width=\myW]{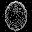}\tabularnewline
		
		3 &
		\includegraphics[width=\myW]{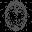} & 
		\includegraphics[width=\myW]{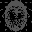} & 
		\includegraphics[width=\myW]{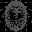} & 
		\includegraphics[width=\myW]{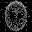} & 
		\includegraphics[width=\myW]{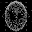} & 
		\includegraphics[width=\myW]{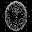} & 
		\includegraphics[width=\myW]{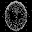} & 
		\includegraphics[width=\myW]{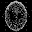} & 
		\includegraphics[width=\myW]{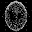} & 
		\includegraphics[width=\myW]{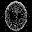}\tabularnewline
		
		10 &
		\includegraphics[width=\myW]{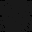} & 
		\includegraphics[width=\myW]{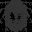} & 
		\includegraphics[width=\myW]{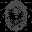} & 
		\includegraphics[width=\myW]{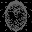} & 
		\includegraphics[width=\myW]{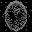} & 
		\includegraphics[width=\myW]{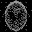} & 
		\includegraphics[width=\myW]{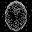} & 
		\includegraphics[width=\myW]{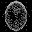} & 
		\includegraphics[width=\myW]{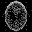} & 
		\includegraphics[width=\myW]{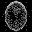}\tabularnewline
		
		50 &
		\includegraphics[width=\myW]{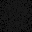} & 
		\includegraphics[width=\myW]{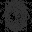} & 
		\includegraphics[width=\myW]{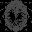} & 
		\includegraphics[width=\myW]{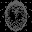} & 
		\includegraphics[width=\myW]{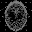} & 
		\includegraphics[width=\myW]{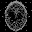} & 
		\includegraphics[width=\myW]{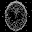} & 
		\includegraphics[width=\myW]{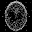} & 
		\includegraphics[width=\myW]{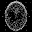} & 
		\includegraphics[width=\myW]{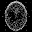}\tabularnewline
		
		100 &
		\includegraphics[width=\myW]{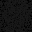} & 
		\includegraphics[width=\myW]{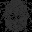} & 
		\includegraphics[width=\myW]{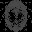} & 
		\includegraphics[width=\myW]{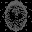} & 
		\includegraphics[width=\myW]{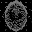} & 
		\includegraphics[width=\myW]{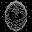} & 
		\includegraphics[width=\myW]{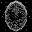} & 
		\includegraphics[width=\myW]{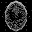} & 
		\includegraphics[width=\myW]{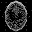} & 
		\includegraphics[width=\myW]{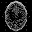}\tabularnewline

	\end{tabular}
	\caption{Visualsing the reconstruction process of Shepp-Logan phantom at different points in the optimisation process (every 10,000 FEs) and for varying number of expanding boxes, with phantom size of $32\times32$ and $6$ projections. } \label{fig:Shepp32x32x6}
\end{figure}

While a single box produces the smallest median reconstruction error (see the first row in Table~\ref{tab:e1e2SheppLogan}, and Table~\ref{tab:e1e2SheppLogan-stat}-left), the result is distant from the original phantom (see the corresponding $e_2$ error and Fig.~\ref{fig:Shepp32x32x6}); 50 boxes, however, produces a better reproduction error, with the reconstructed image closer to the original. Increasing the number of boxes beyond 50 does not statistically significantly alter the $e_2$ outcome (see last row in Table~\ref{tab:e1e2SheppLogan-stat}-right).

\begin{table}[t]
\caption{Median error values for 30 runs on each of five boxing scenarios for $32\times32$ Shepp-Logan, imaged with 6 projections. The smallest $e_1$ and $e_2$ errors are highlighted. } \label{tab:e1e2SheppLogan} \tiny  \scriptsize
\centering
\vspace{2mm} \setlength{\tabcolsep}{5.3pt}
\begin{tabular}{lcc}
	\hline
	Boxes  & $e_{1}$  & $e_{2}$ \tabularnewline
	\hline 
	1  & \cellcolor{black!20} 47222 & 32642 \tabularnewline
	2  & 113234 & 24731 \tabularnewline
	3  & 102552 & 20463 \tabularnewline
	10  & 110200 & 20085 \tabularnewline
	50  & 114391 & \cellcolor{black!20} 19662 \tabularnewline
	100  & 120111 & 19957 \tabularnewline
	\hline 
\end{tabular}
\end{table}

\begin{table}[t]
\caption{Statistical analysis of $e_1$ (left) and $e_2$ (right) for varying boxing scenarios for $32 \times 32$ Shepp Logan with 6 projections. A `1' indicates a Wilcoxon win for the algorithm over 30 runs in the left hand column versus the algorithm in the top row. `0' indicates no significant difference in the algorithms. For the reproduction error, there are no significant differences from boxing scenario 50.} \tiny \scriptsize
\label{tab:e1e2SheppLogan-stat}
\setlength{\tabcolsep}{1.3pt}
\renewcommand\arraystretch{1.12} \centering
$e_1$ \hspace{5cm} $e_2$ 

\begin{tabular}{|l||c|c|c|c|c|c|c|c|c|c|}
	\hline 
	\textbf{Boxes} & 1  & 2  & 3  & 10  & 50  & 100  \tabularnewline
	\hline 
	\hline 
	1 & \cellcolor{black!25} NA & 1  & 1  & 1  & 1  & 1  \tabularnewline 
	\hline  
	2 & 0  & \cellcolor{black!25} NA & 0  & 0  & 0  & 1   \tabularnewline 
	\hline  
	3 & 0  & 1  & \cellcolor{black!25} NA & 1  & 1  & 1   \tabularnewline 
	\hline  
	10 & 0  & 0  & 0  & \cellcolor{black!25} NA & 1  & 1    \tabularnewline 
	\hline  
	50 & 0  & 0  & 0  & 0  & \cellcolor{black!25} NA & 1   \tabularnewline 
	\hline  
	100 & 0  & 0  & 0  & 0  & 0  & \cellcolor{black!25} NA  \tabularnewline 
	\hline  
\end{tabular}
\hspace{3mm}
\begin{tabular}{|l||c|c|c|c|c|c|c|c|c|c|}
	\hline 
	\textbf{Boxes} & 1  & 2  & 3  & 10  & 50  & 100  \tabularnewline
	\hline 
	\hline 
	1 & \cellcolor{black!25} NA & 0  & 0  & 0  & 0  & 0    \tabularnewline 
	\hline  
	2 & 1  & \cellcolor{black!25} NA & 0  & 0  & 0  & 0   \tabularnewline 
	\hline  
	3 & 1  & 1  & \cellcolor{black!25} NA & 0  & 0  & 0    \tabularnewline 
	\hline  
	10 & 1  & 1  & 1  & \cellcolor{black!25} NA & 0  & 0    \tabularnewline 
	\hline  
	50 & 1  & 1  & 1  & 1  & \cellcolor{black!25} NA & 0   \tabularnewline 
	\hline  
	100 & 1  & 1  & 1  & 1  & 0  & \cellcolor{black!25} NA  \tabularnewline 
	\hline  
\end{tabular}
\end{table}

\subsection{Total variation regularisation}

Total variation regularisation is another tool in `denoising' images during the reconstruction process. The successful use of this method depends highly on adjusting the $\mu$ parameter (Eq.~\ref{eq:TV}). To conduct the fine-tuning, a range of values from $\mu=[0,1000]$ are selected and 30 independent trials are run for each $\mu$ value. Table~\ref{tab:mu} presents (a) the median error values for both $e_1$ and $e_2$, along with (b) the statistical analysis of the results. 
Based on the statistical analysis of the reproduction error, the `sweet spot' for the regularisation parameter is between $\mu=[40,65]$, with the best value $\mu=55$ with 19 significant wins out of 25. The regularisation, which is studied here, follows the search space expansion, and serves the similar purpose of smoothing the reconstructed images by balancing the trade off between a smaller $e_1$ error with a more clinically plausible reconstruction. We notice that, as in the boxing scenarios, regularisation degrades the $e_1$ error as shown in Table~\ref{tab:mu} (the best $e_1$ is obtained when $\mu=0$), but improves the overall reconstruction, as seen from its impact on $e_2$ error.

\begin{table}
\caption{Total variation regularisation. (a) $e_1$ and $e_2$ median error values for varying $\mu$, along with (b) the count of statistically significant wins for each $\mu$.}
\label{tab:mu}
\setlength{\tabcolsep}{5.3pt}
\renewcommand\arraystretch{1.12} \centering \tiny \scriptsize

\begin{tabular}{|r||rr||rr|}
	\hline 
	& \multicolumn{2}{c||}{ (a) Error values} & \multicolumn{2}{p{20mm}|}{(b) Statistically significant cases}\tabularnewline
	\hline 
	\textbf{$\mu$}  & $e_{1}$  & $e_{2}$  & \multicolumn{1}{r}{$e_{1}$ } & $e_{2}$ \tabularnewline
	\hline 
	\hline 
	0  & \cellcolor{black!10}118549  & 19833  & \hspace{3mm} \cellcolor{black!20} 24   & 1\tabularnewline
	1  & 176868  & 19457  & 23  & 2\tabularnewline
	5  & 401477  & 18274  & 22  & 3\tabularnewline
	10  & 642329  & 17213  & 21  & 4\tabularnewline
	15  & 846614  & 16379  & 20  & 6\tabularnewline
	20  & 1033289  & 15807  & 19  & 8\tabularnewline
	25  & 1216424  & 15353  & 18  & 9\tabularnewline
	30  & 1386386  & 14972  & 17  & 12\tabularnewline
	35  & 1556795  & 14734  & 16  & 16\tabularnewline
	40  & 1719532  & \cellcolor{black!10}14580  & 15  & 18\tabularnewline
	45  & 1889283  & \cellcolor{black!10}14630  & 14  & 17\tabularnewline
	50  & 2052300  & \cellcolor{black!10}14625  & 13  & 18\tabularnewline
	55  & 2187202  & \cellcolor{black!10}14650  & 12  & \cellcolor{black!20} 19\tabularnewline
	60  & 2352773  & \cellcolor{black!10}14672  & 11  & 16\tabularnewline
	65  & 2507530  & \cellcolor{black!10}14577  & 10  & 18\tabularnewline
	70  & 2665960  & 14762  & 9  & 15\tabularnewline
	75  & 2813415  & 14682  & 8  & 14\tabularnewline
	80  & 2957201  & 14861  & 7  & 13\tabularnewline
	85  & 3098517  & 14906  & 6  & 13\tabularnewline
	90  & 3249197  & 15097  & 5  & 12\tabularnewline
	95  & 3410860  & 15243  & 4  & 9\tabularnewline
	100  & 3539566  & 15306  & 3  & 9\tabularnewline
	150  & 4899699  & 16182  & 2  & 6\tabularnewline
	200  & 6174531  & 17193  & 1  & 4\tabularnewline
	1000  & 24795852  & 25632  & 0  & 0\tabularnewline
	\hline 
\end{tabular}
\end{table}

\subsection{Comparison with SHADE-ILS algorithm}

The standard version of the optimisers used in this work are designed for lower dimensional problems. Large-scale global optimisation (LSGO), which deals with a great number of variables, has been gaining increasing popularity in recent years. One such method, SHADE with iterative local search or SHADE-ILS~\cite{molina2018_SHADE-ILS}, which has won the CEC’2018 LSGO competition, hybridises SHADE and two local search methods: MTS-L1~\cite{tseng2008_MTS-L1} and L-BFGS-B~\cite{zhu1997_L-BFGS-B}. The performance of this algorithm along with its use of local search methods, provides a good basis for comparison with the proposed swarm technique. In this comparison, the optimised DFO for TR, with neither the search space expansion nor total variation regularisation (DFO-TR), is compared against DFO-TR with search space expansion and total variation regularisation (DFO-TR-$\mu$), the best performing TR toolbox algorithm (SIRT), and SHADE-ILS.

Wilcoxon significance results for 30 runs on Shepp-Logan phantom are given in Table~\ref{tab:SHADE-ILS}.
Table~\ref{tab:SHADE-ILS}-left compares algorithm pairs when rated according to the reconstruction error, $e_1$. Unsurprisingly, SIRT produces the best reconstruction, $e_1$ (followed by SHADE-ILS); while DFO-TR does not offer a statistically significantly difference with SHADE-ILS in image reproduction, $e_2$, DFO-TR-$\mu$ which benefits from the dual regularisation methods, produces statistically significant smaller reproduction errors (Table~\ref{tab:SHADE-ILS}-right).

Fig.~\ref{fig:finalComparison} shows the visual reconstruction of the Shepp-Logan phantom using the four aforementioned techniques.
The figure shows the presence of artefacts in SIRT, and salt-and-pepper noise in both DFO-TR and SHADE-ILS. Under this highly undersampled scenario, DFO-TR-$\mu$ presents a respectable reconstruction of the phantom, albeit with yet some distance from a zero reproduction error.

\begin{table}[t]
\caption{Comparison with SHADE-ILS. A `1' indicates a Wilcoxon win for the algorithm over 30 runs in the left hand column versus the algorithm in the top row. `0' indicates no significant difference in the algorithms.}
\label{tab:SHADE-ILS} \tiny \scriptsize
\setlength{\tabcolsep}{1.3pt}
\renewcommand\arraystretch{1.25} 
\centering
\begin{tabular}{|l||c|c|c|c|c|c|}
\hline 
\textbf{$e_1$} & DFO-TR &  DFO-TR-$\mu$ & SIRT & SHADE-ILS \tabularnewline
\hline
\hline
DFO-TR & \cellcolor{black!25} NA & 1  & 0  & 0  \tabularnewline 
\hline  
DFO-TR-$\mu$ & 0  & \cellcolor{black!25} NA & 0  & 0  \tabularnewline 
\hline  
SIRT & 1  & 1  & \cellcolor{black!25} NA & 1  \tabularnewline 
\hline  
SHADE-ILS & 1  & 1  & 0  & \cellcolor{black!25} NA  \tabularnewline 
\hline    
\end{tabular}
\hspace{5mm}
\begin{tabular}{|l||c|c|c|c|c|c|}
\hline 
\textbf{$e_2$} & DFO-TR  &  DFO-TR-$\mu$  & SIRT &SHADE-ILS \tabularnewline
\hline 
\hline 
DFO-TR & \cellcolor{black!25} NA & 0  & 1  & 0  \tabularnewline 
\hline  
DFO-TR-$\mu$ & 1  & \cellcolor{black!25} NA & 1  & 1  \tabularnewline 
\hline  
SIRT & 0  & 0  & \cellcolor{black!25} NA & 0  \tabularnewline 
\hline  
SHADE-ILS & 0  & 0  & 1  & \cellcolor{black!25} NA  \tabularnewline 
\hline   
\end{tabular}
\vspace{3mm}
\end{table}

\begin{figure}
\centering
\newcommand{\myW}{0.16\linewidth} 
\setlength{\tabcolsep}{1pt}
\begin{tabular}{cccccc} 
Phantom & DFO-TR & DFO-TR-$\mu$  & SIRT  & SHADE-ILS \tabularnewline %
\includegraphics[width=\myW]{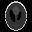} &
\includegraphics[width=\myW]{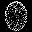} & 
\includegraphics[width=\myW]{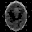} & 
\includegraphics[width=\myW]{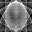} & 
\includegraphics[width=\myW]{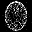} \tabularnewline

\end{tabular}
\caption{Reconstructing Shepp-Logan by the swarm optimiser without the smoothing methods (DFO-TR), the presented method with dual regularisation of search space expansion and the total variation (DFO-TR-$\mu$), the best standard TR toolbox algorithm (SIRT), and SHADE-ILS.} \label{fig:finalComparison}
\end{figure}

\section{Conclusion}

This work investigates two regularisation methods in the context of tomographic reconstruction when dealing with highly undersampled data in few-view scenarios.
These methods -- search space expansion and total variation -- are sequentially fine-tuned and adapted to the reconstructing algorithm. 
Integrating these methods results in the worsening of the reconstruction errors, which was expected; however, the presented pipeline, and the use of the updated objective function, result in lower reproduction errors, which demonstrates a more faithful reconstruction to the ground truth than the standard toolbox algorithms and a high-dimensional optimiser.
Among the subjects of future research is 
reformulating the problem into a multiobjective problem
as well as hybridising classical toolbox techniques (with their signature feature of low reconstruction error) and the presented swarm optimisation method, benefiting from dual regularisation (with low reproduction errors) which could produce a powerful algorithm capable of fast noise- and artefacts-free reconstructions in the few-view regime.

\bibliographystyle{plain}
\bibliography{mybibliography}

\end{document}